\documentclass[sigconf]{acmart}

\usepackage{amsthm}
\usepackage{amsmath}
\usepackage{subcaption}
\usepackage{paralist}
\usepackage{cases}
\usepackage{booktabs}
\usepackage{threeparttable}
\usepackage{epstopdf}
\usepackage{upgreek}
\usepackage{endnotes}
\usepackage{etoolbox}
\usepackage{algpseudocode}
\usepackage{xspace}
\usepackage{array}
\usepackage{enumitem}
\usepackage{balance}
\usepackage{multirow}
\usepackage{xcolor}
\usepackage{graphicx}
\usepackage{wrapfig}
\usepackage{enumitem}
\usepackage{tabularx}
\usepackage{float}
\usepackage{svg}
\usepackage{subcaption}
\usepackage{flushend}
\usepackage[ruled,linesnumbered]{algorithm2e}
\usepackage{stfloats}
\usepackage{afterpage}
\AtBeginDocument{%
  }

\newcommand{\ie}{\emph{i.e.,}\xspace}

\newcommand{\TODO}[1]{{\color{red}TODO: {#1}}}

\newcommand{\eat}[1]{}

\setcopyright{acmlicensed}
\copyrightyear{2018}
\acmYear{2018}
\acmDOI{XXXXXXX.XXXXXXX}
\acmConference[Conference acronym 'XX]{Make sure to enter the correct
  conference title from your rights confirmation email}{June 03--05,
  2018}{Woodstock, NY}
\acmISBN{978-1-4503-XXXX-X/2018/06}

\usepackage{amsthm}
\usepackage{multirow} 
\newtheorem{definition}{Definition}
\newtheorem{problem}{Problem}

\begin{document}

\title{CoLLMLight: Cooperative Large Language Model Agents for Network-Wide Traffic Signal Control}


\author{Zirui Yuan}
\affiliation{\country{}
  \institution{The Hong Kong University of Science and Technology (Guangzhou)}}
\email{zyuan779@connect.hkust-gz.edu.cn}

\author{Siqi Lai}
\affiliation{\country{}
  \institution{The Hong Kong University of Science and Technology (Guangzhou)}}
\email{slai125@connect.hkust-gz.edu.cn}

\author{Hao Liu}
\authornote{Corresponding author}
\affiliation{\country{}
  \institution{The Hong Kong University of Science and Technology (Guangzhou)}}
\email{liuh@ust.hk}







\renewcommand{\shortauthors}{Trovato et al.}

\begin{abstract}
Traffic Signal Control (TSC) plays a critical role in urban traffic management by optimizing traffic flow and mitigating congestion. While Large Language Models (LLMs) have recently emerged as promising tools for TSC due to their exceptional problem-solving and generalization capabilities, existing approaches fail to address the essential need for inter-agent coordination, limiting their effectiveness in achieving network-wide optimization. To bridge this gap, we propose \textbf{CoLLMLight}, a cooperative LLM agent framework for TSC. 
Specifically, we first construct a structured spatiotemporal graph to capture real-time traffic dynamics and spatial relationships among neighboring intersections, enabling the LLM to reason about complex traffic interactions.
Moreover, we introduce a complexity-aware reasoning mechanism that dynamically adapts reasoning depth based on real-time traffic conditions, ensuring optimal computational efficiency without sacrificing decision quality.
Besides, we propose a fine-tuning strategy that leverages iterative simulation-driven data collection and environmental feedback to build a lightweight LLM tailored for cooperative TSC. 
Extensive experiments on both synthetic and real-world datasets demonstrate that CoLLMLight outperforms state-of-the-art methods in diverse traffic scenarios, showcasing its effectiveness, scalability, and robustness. Our code is available at \url{https://github.com/usail-hkust/CoLLMLight}.
\end{abstract}

\eat{

}


\maketitle

\section{Introduction}

Traffic congestion has become a critical challenge, significantly affecting both society and the urban environment. With the rapid acceleration of urban migration, city populations are expanding, further exacerbating this issue. Traffic Signal Control (TSC) is vital in optimizing traffic flow and improving road safety \cite{wu2023transformerlight, zhang2024irregular, wei2019survey}. Over the past decades, numerous studies have explored both transportation-based and data-driven approaches to TSC. However, effectively managing traffic signals in complex and dynamic urban road networks remains a formidable challenge.

Transportation-based methods primarily rely on heuristic algorithms that dynamically adjust signals based on real-time lane-level traffic conditions \cite{roess2004traffic, little1981maxband, cools2013self, hunt1982scoot, koonce2008traffic, lowrie1990scats, varaiya2013max}. However, these approaches often require extensive manual design and struggle to generalize across complex, dynamic traffic patterns \cite{wei2019survey}. In contrast, data-driven methods leverage reinforcement learning (RL) to learn optimal control policies through continuous environmental interactions, capturing multi-level embeddings of traffic dynamics~\cite{oroojlooy2020attendlight, chen2020toward, wei2019presslight, liang2022oam, wu2023transformerlight}. To further enhance coordination across intersections and achieve road network-wide optimization, recent studies~\cite{devailly2021ig, wei2019colight, lou2022meta, wu2021dynstgat, yu2021macar} integrate graph neural networks (GNNs)~\cite{bohmer2020deep} to efficiently aggregate spatial information from neighboring intersections. These RL-based approaches have achieved state-of-the-art performance by leveraging collaborative traffic signal control. However, RL-based methods face significant generalization challenges, particularly in adapting to unseen or highly variable traffic conditions, due to their reliance on limited training data and data-centric optimization processes \cite{lai2023large}.

\begin{figure}[t]
    \centering
    \begin{subfigure}[b]{0.55\linewidth}
        \centering
        \includegraphics[width=\textwidth]{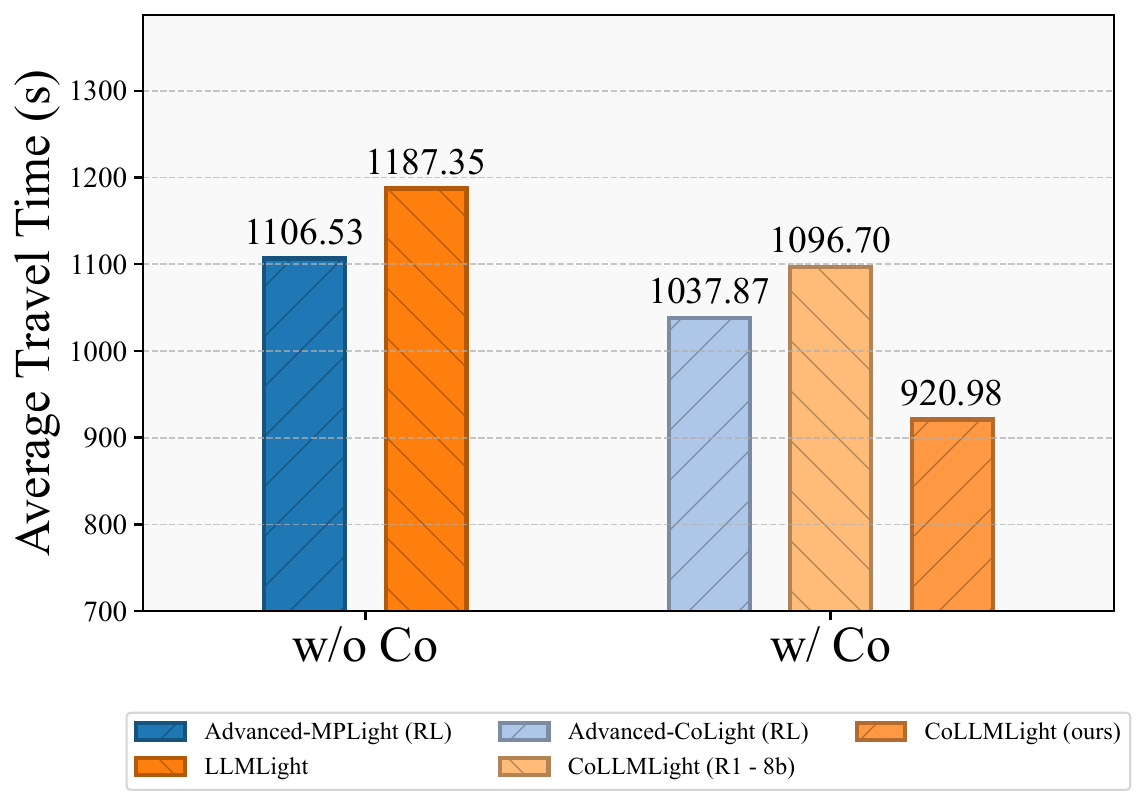}
        \caption{Perf. w/ or w/o Cooperation}
        \label{fig:fig1a}
    \end{subfigure}
    \hfill
    \begin{subfigure}[b]{0.4\linewidth}
        \centering
        \includegraphics[width=\textwidth]{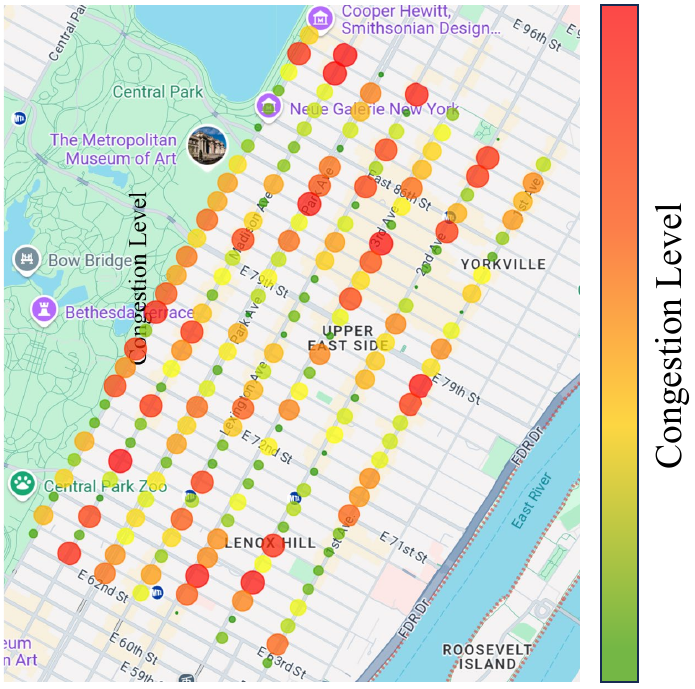}
        \caption{Congestion Distribution in New York Dataset}
        \label{fig:fig2b}
    \end{subfigure}
    \caption{(a): Both RL-based and LLM-based methods can benefit from cross-intersection collaboration. (b): Congestion levels vary significantly across different locations.}
    \label{fig:fig1}
\end{figure}

The advent of Large Language Models (LLMs) has introduced a paradigm shift across multiple domains, showcasing exceptional problem-solving and generalization capabilities. Recent studies have demonstrated their potential in TSC, highlighting their ability to tackle complex tasks with human-like reasoning \cite{lai2023large, feng2024citybench}. However, existing LLM-based approaches are limited by their focus on localized decision-making, neglecting the critical aspect of cross-intersection collaboration necessary for global traffic optimization. As illustrated in Figure \ref{fig:fig1}(a), introducing neighborhood information significantly enhances the performance of both RL-based and LLM-based TSC approaches, emphasizing the importance of cross-intersection collaboration in effective traffic management. Despite this, enabling LLMs to reason about intricate traffic patterns across multiple neighboring intersections presents two key challenges.

The first challenge lies in enabling LLM agents to comprehend the intricate traffic interactions among neighboring intersections. This requires, first, identifying spatial relationships, such as lane connectivity and inter-dependencies between intersections. Second, the agents need to reason about dynamic traffic interactions, including real-time congestion propagation and queue spillbacks. A critical aspect is the ability to predict how their local decisions will influence not only the assigned intersection but also adjacent ones, ensuring that coordination leads to network-wide improvements. This presents a complex cooperation problem, where each agent must simultaneously model both spatiotemporal dependencies and anticipate cascading effects.

Achieving this complex coordination requires a nuanced reasoning process to ensure accurate decision-making. One approach to this is inference-time scaling \cite{snell2024scaling, guo2025deepseek}, which involves extending the length of the Chain-of-Thought (CoT) \cite{wei2022chain}, allowing models to perform deeper, more structured reasoning. However, real-world traffic conditions vary significantly across locations, as illustrated in Figure \ref{fig:fig1}(b). Some intersections experience light traffic with minimal vehicle presence, making signal optimization relatively straightforward. In contrast, heavily congested intersections require more sophisticated reasoning, as optimal signal selection depends on complex interactions with neighboring traffic flows. Applying the same reasoning depth to all intersections is inefficient: simpler cases do not benefit from extensive computation, while congested areas demand more nuanced analysis. A uniform approach not only wastes computational resources but is also impractical for real-time decision-making at scale. Thus, achieving a balance between computational efficiency and decision-making accuracy is another critical challenge.

To tackle these challenges, we propose CoLLMLight, a cooperative LLM agent framework tailored for TSC. To enable the LLM to comprehend intricate traffic interactions, we first collect real-time traffic state information from neighboring intersections over a recent time window. Then, a structured spatiotemporal graph is constructed to capture both spatial connectivity and temporal traffic dynamics. Leveraging this graph, the LLM predicts future traffic states under different signal configurations, evaluates their network-wide impacts, and selects the optimal signal to improve overall traffic conditions.

Furthermore, to ensure both efficiency and effectiveness, we introduce a complexity-aware reasoning mechanism that classifies decision-making complexity into three levels—ranging from lightly to highly congested intersections—and adapts reasoning strategies accordingly. The agent dynamically adjusts its reasoning depth based on real-time traffic conditions, optimizing computational resource allocation without sacrificing decision quality. 

To further enhance performance, we propose a fine-tuning strategy for building a lightweight LLM tailored for cooperative traffic signal control. Through an iterative simulation-driven data collection process, we synthesize task-specific reasoning chains to improve traffic pattern comprehension. We then employ GPT-4o to evaluate simulation outputs and generate high-quality pseudo labels for reasoning complexity identification, future traffic state prediction, and signal selection. Additionally, a policy refinement process with environmental feedback iteratively improves decision-making, ensuring maximized network-wide traffic efficiency.

\eat{To address these challenges, we propose CoLLMLight, an effective and efficient cooperative LLM agent specifically designed for TSC. To enable globally optimal decision-making, we introduce \textit{Decision Making with spatiotemporal Deduction (DMSTD)}. Specifically, we model the current and historical traffic states across multiple intersections as a spatiotemporal graph. The agent then deduces the future traffic states that each signal will generate. By comparing these future states, the agent gains insight into the impact of each signal and can determine the optimal signal that leads to improved future conditions. 
Furthermore, to ensure both efficiency and effectiveness, we propose \textit{Coordination-Aware Hybrid Reasoning}. Specifically, we classify the coordination complexity of TSC into three levels and design two additional, simpler reasoning strategies. By integrating these strategies with DMSTD, the agents can dynamically select a cost-effective reasoning approach based on current traffic conditions. The entire methodology is refined through reasoning tuning with environmental feedback.}

Our contributions are summarized as follows: 
1) We propose \textbf{CoLLMLight}, a cooperative LLM agent framework for network-wide traffic signal control. To our knowledge, this is the first work to integrate LLM cooperation into TSC. 
2) We design a complexity-aware reasoning mechanism that dynamically adapts reasoning modes based on real-time traffic conditions, ensuring an optimal balance between computational efficiency and performance. 
3) We propose an LLM fine-tuning method for cooperative TSC, which learns traffic dynamics through iterative simulation and environmental feedback, leading to significant improvements in decision-making and efficiency. 
4) Extensive evaluations on both synthetic and real-world datasets validate the effectiveness and superiority of CoLLMLight in diverse traffic scenarios.

\section{Problem Statement}  
In this section, we introduce key concepts related to traffic signal control and the formal problem statement. A more detailed illustration of the most used setting of the intersection, lanes, and signal phases are in Appendix \ref{sec:tsc_setting}

\begin{definition}[Road Network]  
The road network is a directed graph connected by intersections $\mathcal{V}$ and lanes $\mathcal{L}$. Lanes can be categorized into three types:
1) \textit{go-through} lanes ($\mathcal{L}_{go}$),  
2) \textit{left-turn} lanes ($\mathcal{L}_{left}$),  
3) \textit{right-turn} lanes ($\mathcal{L}_{right}$).  
These lanes are interconnected with their neighboring intersections.
\end{definition}

\begin{definition}[Traffic Signals]  
At each signal-switching time step, the agent assigned to the intersection selects a signal from the predefined signal set $\mathcal{A} = \{a_1, \ldots, a_m\}$. The traffic signal is represented as $a = \text{set}(\mathcal{L}_{allow})$, where $\mathcal{L}_{allow}$ is a group of allowed-to-go lanes without conflicting movements (\ie a green light for $\mathcal{L}_{allow}$ and a red light for others).
\end{definition}

\begin{problem}[Multi-Agent LLM for Network-Wide Traffic Signal Control]  
Consider a road network with multiple intersections, each intersection is controlled by an LLM agent with a control policy $\pi$. At each signal-switching timestep $t$, each agent receives:  
1) traffic observations from both assigned and neighboring intersections $\mathcal{O}_t$;  
2) spatial relations among intersections $\mathcal{G}$;  
3) historical traffic interactions $\mathcal{T}_t$;  
4) relative task descriptions $\mathcal{D}$.  
Based on these inputs, the agent reasons the optimal traffic signal action $a_t$ from the action space $\mathcal{A}$, with the goals of maximizing the network-wide traffic efficiency:
\begin{equation}  
    a_t = \pi\left([\mathcal{O}_t, \mathcal{G}, \mathcal{T}_t], \mathcal{D}, \mathcal{A}\right).   
\end{equation}   
\end{problem}

\begin{figure*}[htbp]  
    \centering  
    \includegraphics[width=\textwidth]{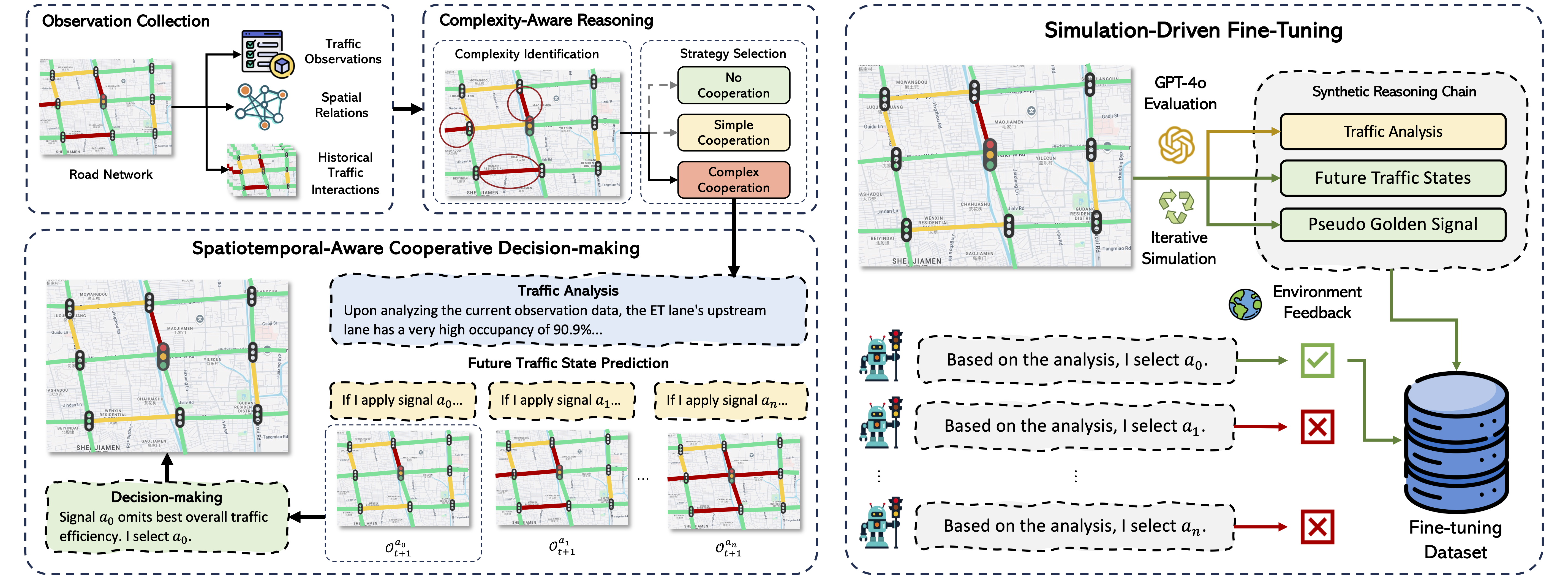}  
    \caption{The overview of CoLLMLight framework.}  
    \label{fig:main_figure}  
\end{figure*}

\section{Cooperative LLM Agent Framework}
We present the framework of CoLLMLight in Figure \ref{fig:main_figure}, including:
1) \textit{Spatiotemporal-aware cooperative decision-making}: It enables the LLM to understand and leverage spatiotemporal information, allowing it to capture intricate traffic interactions and make accurate, coordinated decisions.
2) \textit{Complexity-aware reasoning}: By evaluating the current traffic conditions, it dynamically selects the most appropriate reasoning mode, ensuring that computational resources are efficiently allocated without sacrificing decision quality.
3) \textit{Simulation-driven fine-tuning}: Through iterative simulation, this component generates task-specific data that is used to fine-tune the LLM, thereby enhancing its decision-making accuracy and efficiency in cooperative TSC scenarios.

\eat{
We present the framework of CoLLMLight in Figure \ref{fig:main_figure}, including:
1) \textit{Cooperative Observation Construction}: This step facilitates real-time communication among multiple agents to aggregate diverse traffic features, ensuring a comprehensive understanding of the traffic environment;
2) \textit{Coordination-Aware Reasoning}: This stage evaluates the complexity of current traffic conditions and dynamically selects the most suitable reasoning mode, balancing decision quality and computational efficiency to determine the optimal traffic signal.
\TODO{revise this after refine main figure}
}

\subsection{Spatiotemporal-Aware Cooperative Decision-Making}
\label{sec:dmstd}
Complex traffic coordination scenarios require the LLM agent to infer traffic states not only at its assigned intersection but also at neighboring intersections. This inference is based on the spatial dependencies between intersections and the temporal dynamics of congestion propagation across the network. We propose a spatiotemporal-aware cooperative decision-making mechanism that enables the LLM agent to effectively integrate both spatial and temporal information, thereby optimizing traffic flow across the road network. 
\eat{This approach allows the agent to consider the broader context of traffic conditions, making more informed and coordinated decisions.}

\eat{To enable the LLM agent to comprehensively assess the impacts of signal changes, we proposed spatiotemporal deduction step within the decision-making process. This step requires the agent to infer the traffic states of both its own intersection and neighboring intersections for the next signal period based on the current spatiotemporal information associated with each signal. Subsequently, the agent selects the signal that optimally leads to the most efficient traffic states for the upcoming period.}

\subsubsection{Observation Collection}
We begin by formulating lane-level observations at the current timestep $t$ as:
\begin{align}
    \mathbf{o}_t = \left[ n^{queue}, n^{move}, \tau, \rho \right],
\end{align}
where $n^{queue}$ represents the number of vehicles in the queue, $n^{move}$ denotes the number of vehicles currently moving, $\tau$ indicates the average waiting time, and $\rho \in [0, 1]$ reflects the traffic occupancy of the lane. These lane-level observations are then aggregated to construct both local and neighboring observations at the assigned and nearby intersections as $\mathcal{O}_t = \{\mathbf{o}_t | l \in \mathcal{L}\},$
where $\mathcal{L}$ represents the set of lanes connected to the agent's assigned intersection and its neighboring intersections.

\subsubsection{Spatiotemporal Modeling}
In the road network topology, lanes exhibit directional connectivity across intersections. To represent these spatial relationships, we construct a directed subgraph for each intersection, denoted as $\mathcal{G} = (\mathcal{V}, \mathcal{L})$, where $\mathcal{V}$ is the set of intersections (including both the assigned and neighboring intersections), and $\mathcal{L}$ denotes the set of adjacent lanes connecting these intersections.

To capture the temporal pattern of congestion propagation, we collect historical traffic interactions over fixed time windows $\Delta t$, which include past traffic observations with corresponding activated signal configurations:
\begin{align}
    \mathcal{T}_t = \{(\mathcal{O}_{t_i}, \mathbf{a}_{t_i})|t-\Delta t < t_i < t\},
\end{align}
where $\mathbf{a}_{t_i}$ represents the signal configurations at both the assigned and neighboring intersections at time $t_i$.

\subsubsection{Cooperative Decision-Making}\label{subsubsec:cooperative_decision_making}
To enable the LLM to understand our constructed observations and spatiotemporal features, we generate a human-readable prompt that combines current traffic observations $\mathcal{O}_t$, the spatial relation graph $\mathcal{G}$, historical traffic interactions $\mathcal{T}_t$, and task instructions $\mathcal{D}$ (e.g., preventing vehicle release into congested downstream lanes or dynamically adjusting lane usage in response to upstream congestion). To scale the chain-of-thought (CoT) process, the LLM follows a detailed three-step decision-making process: First, it analyzes the current traffic conditions at both the assigned and neighboring intersections. Next, it predicts neighboring future traffic states under different signal configurations. Finally, based on these predictions, it selects the signal configuration that is most likely to optimize network-wide traffic flow. Formally, this process is represented as:
\begin{align}
    & \mathbf{X} = \text{Prompt}(\mathcal{O}_t, \mathcal{G}, \mathcal{T}_t, \mathcal{D}), \\
    & \mathbf{Y}_\text{ana}, \{\hat{\mathcal{O}}^a_{t+1} | a \in \mathcal{A}\}, \hat{a} = f_{\text{LLM}}(\mathbf{X}),
\end{align}
where $\mathbf{X}$ is the verbalized LLM input, $\mathbf{Y}_\text{ana}$ is the analysis on the current traffic condition, $\hat{\mathcal{O}}^a_{t+1}$ represents the predicted traffic states for each signal configuration $a$ at the next timestep, and $\hat{a}$ is the estimated optimal signal.

\begin{figure}[t]  
    \centering  
    \includegraphics[width=\linewidth]{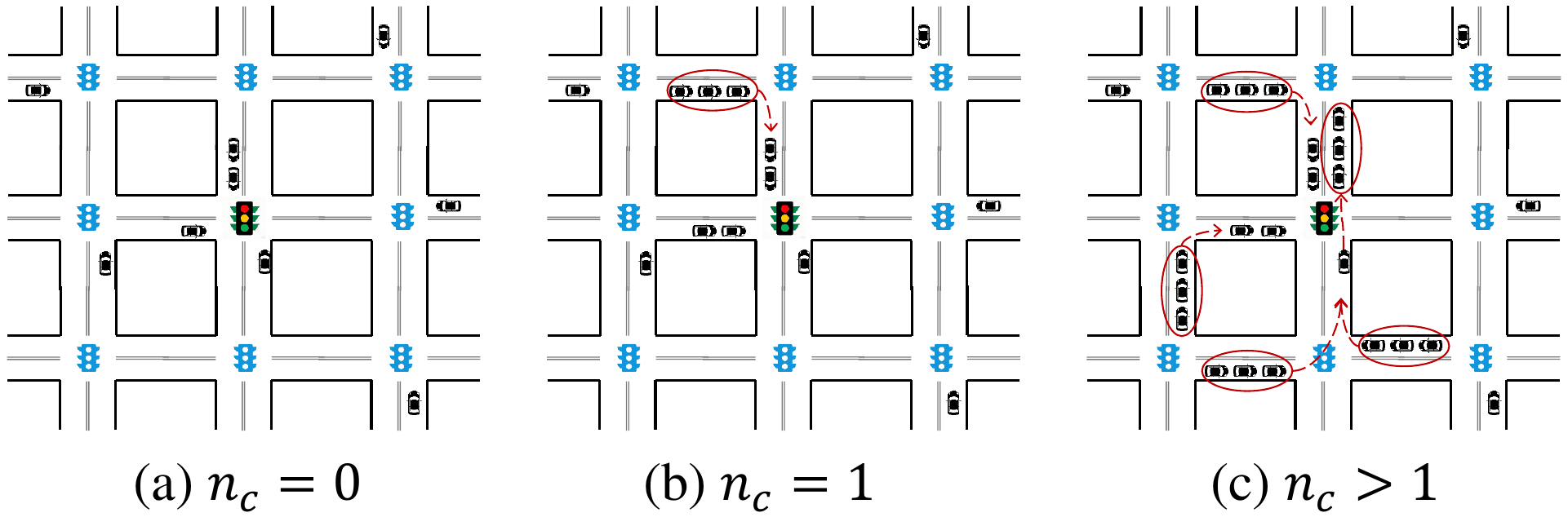}  
    \caption{An illustration of traffic scenarios with varying coordination complexity. Critical lanes are circled in red. (a) $n_c=0$ (No cooperation): Signals can determined locally. (b) $n_c=1$ (Simple cooperation): The agent needs to consider the interaction with one critical lane. (c) $n_c>1$ (Complex Cooperation): The agent needs to consider interactions with multiple critical lanes.}
    \label{fig:coordination_complexity}
\end{figure}

\subsection{Complexity-Aware Reasoning}
While a spatiotemporal-aware decision-making process significantly improves accuracy by capturing complex traffic dependencies among intersections, it often incurs substantial inference time due to the need for generating long token sequences. Applying such computationally intensive reasoning uniformly across all scenarios can be impractical, as it may compromise real-time responsiveness in dynamic traffic environments. To address this, we propose a complexity-aware reasoning approach, enabling the LLM to automatically assess the complexity of decision-making at each intersection and adaptively select the most cost-efficient reasoning strategy without sacrificing performance.

We first classify the reasoning complexity at an intersection based on a congestion risk coefficient $ n_c $, which is the number of critical neighboring lanes that are either congested or at risk of congestion. As illustrated in Figure \ref{fig:coordination_complexity}: 

\noindent 1) $ n_c = 0 $ (\textit{No Cooperation}): When no critical neighboring lanes are congested, the intersection operates independently, requiring no complex cooperative decision-making for signal optimization. In this case, both traffic condition analysis and future traffic state prediction are omitted:
\begin{align}
    \hat{a} = f^{\text{no-coop}}_{\text{LLM}}(\mathbf{X}).
\end{align}
2) $ n_c = 1 $ (\textit{Simple Cooperation}): When exactly one neighboring lane exhibits high traffic occupancy, the agent must analyze local spatiotemporal dependencies between adjacent intersections. However, since the congestion is limited to a single neighboring lane, the future traffic state prediction process is excluded.
\begin{align}
    & \mathbf{Y}_\text{ana}, \hat{a} = f^{\text{simple}}_{\text{LLM}}(\mathbf{X}).
\end{align}
3) $ n_c > 1 $ (\textit{Complex Cooperation}): When multiple neighboring lanes are highly congested, the agent must account for more extensive spatiotemporal dependencies, including multi-hop effects across intersections. In this scenario, the complete cooperative decision-making process is applied:
\begin{align}
    & \mathbf{Y}_\text{ana}, \{\hat{\mathcal{O}}^a_{t+1} | a \in \mathcal{A}\}, \hat{a} = f^{\text{complex}}_{\text{LLM}}(\mathbf{X}).
\end{align}
We instruct the LLM to identify the congestion risk coefficient $n_c$ and dynamically select the corresponding reasoning strategy.

\eat{Let $ n_c $ denote the number of critical neighboring lanes that are either congested or at risk of congestion. We classify traffic conditions into three types based on the value of $ n_c $. 

As illustrated in Figure \ref{fig:coordination_complexity}(a), when $ n_c = 0 $, there are no critical neighboring lanes, indicating that the surrounding areas are in a healthy, low-occupancy state, which we refer to as the \textit{no-coordination} scenario. In this scenario, the intersection does not need to coordinate with other intersections and can focus solely on the impact of its own signal.  

When $ n_c = 1 $, the agent must account for the interaction with a single neighboring lane. As shown in Figure \ref{fig:coordination_complexity}(b), there is one upstream lane with high occupancy, and the agent needs to determine whether it can cooperate with this lane to preemptively release vehicles and mitigate congestion, a relatively straightforward task. We classify this condition as \textit{simple coordination}.  
When $ n_c > 1 $, the agent must navigate interactions with multiple neighboring lanes, which may involve conflicting coordination requirements and present a complex coordination problem. As demonstrated in Figure \ref{fig:coordination_complexity}(c), the agent needs to analyze the signal's impact on all critical lanes and balance trade-offs among various intersections.}





\subsection{Simulation-Driven Fine-Tuning}
To further enhance the reasoning capability of LLM in spatiotemporal-aware cooperation, we propose a fine-tuning method that incorporates iterative simulation-based data collection. Our approach consists of two stages: 1) reasoning chain optimization and 2) policy refinement with environmental feedback.

\eat{We propose a reasoning tuning scheme to optimize the spatiotemporal deduction and hybrid reasoning processes. This scheme consists of three stages: 1) reasoning trajectory synthesis, 2) reasoning tuning, and 3) refinement through environmental feedback.}

\subsubsection{Reasoning Chain Optimization}
We first utilize the traffic simulator CityFlow \cite{zhang2019cityflow} with a synthesized traffic flow dataset to generate diverse traffic scenarios spanning various congestion levels. We follow the required LLM outputs specified in Section \ref{subsubsec:cooperative_decision_making}, which include congestion risk $n_c$, traffic analysis $\mathbf{Y}_{\text{ana}}$, future traffic states $\{\mathcal{O}^a_{t+1} \mid a \in \mathcal{A}\}$, and the optimal signal configuration $a^*$ for activation. 

To derive congestion risk and traffic analysis, we utilize GPT-4o to generate a structured summary of traffic conditions at both the assigned intersection and its neighboring intersections: 
\begin{equation}
    n_c, \mathbf{Y}_{\text{ana}} = f_{\text{GPT-4o}}(\mathcal{O}_t).
\end{equation}
To estimate future traffic states, we roll out the simulation under different signal configurations and collect the corresponding next-step traffic states:  
\begin{equation}
\{\mathcal{O}^a_{t+1} \mid a \in \mathcal{A}\} = \{f_{\text{sim}}(\mathcal{O}_t, a) \mid a \in \mathcal{A}\},
\end{equation} 
where $f_{\text{sim}}$ denotes the simulator's state transition function.
To determine the pseudo-golden signal $a^*$, we simulate various signal actions and evaluate their impact on traffic flow. Specifically, we select the signal configuration that minimizes the total number of queued vehicles over a five-timestep horizon in the neighboring intersections:  
\begin{equation}
a^* = \mathop{\text{argmin}}_{a \in \mathcal{A}} f_{\text{queue}}(\mathcal{O}_t, a),
\end{equation} 
where $f_{\text{queue}}$ represents the simulator's queue length measurement function. This pseudo-golden signal serves as an optimal supervisory signal for training.

Given the required LLM outputs for each reasoning strategy employed, we synthesize the corresponding reasoning chain by a structured summarization of GPT-4o. To optimize the LLM, we apply supervised fine-tuning, minimizing the negative log-likelihood of generating the target reasoning chain:
\begin{align}  
    \text{loss} = -\sum_{w=1}^{|Y|} \log P_{\pi}(y_w | \mathbf{X}, \mathbf{Y}_{<w}),  
\end{align}  
where $\mathbf{Y}$ is the synthesized reasoning chain, $P_\pi(y_w | \mathbf{X}, \mathbf{Y}_{<w})$ represents the probability of generating the token $y_w$ given the prompt $\mathbf{X}$ and the preceding tokens in the reasoning chain $\mathbf{Y}_{<w}$.

\subsubsection{Refinement with Environment Feedback}
After optimizing the reasoning chain, the LLM can effectively follow our constructed reasoning process. However, its decision-making remains suboptimal due to potential hallucinations in traffic conditions and misinterpretations of spatiotemporal dependencies. To address this, we propose an iterative refinement process incorporating environmental feedback for further optimization.

Specifically, we simulate different traffic signal configurations based on the LLM's decision-making and evaluate their effectiveness using an environment feedback function, $Q$. A higher feedback value indicates a more favorable signal configuration under the given traffic conditions. We then identify the reasoning chain associated with the signal configuration that yields the highest environmental feedback, which will be designated as a golden pseudo reasoning chain for further fine-tuning:
\begin{align}
    \big\{\mathbf{Y}|\hat{a} = \mathop{\text{argmax}} \limits_{a\in \mathcal{A}} Q(\mathcal{O}_t, a)\big\}^T_{t=0},
\end{align}
where $T$ represents the time window over which environmental feedback is collected. The environment feedback function $Q$ is defined as the inverse of the overall queue length at neighboring intersections after a five-timestep duration, ensuring that lower congestion is rewarded with higher feedback values.

\eat{After reasoning tuning, the LLM can effectively follow the reasoning trajectory. However, its signal selection policy remains suboptimal. To better align with the long-term objective of globally optimized traffic signal control, we propose an iterative refinement process that leverages environmental feedback. Specifically, we further sample new traffic scenarios from the environment and instruct the fine-tuned LLM multiple times until it generates a trajectory that leads to a pseudo-golden signal $\tilde{a}$. These new trajectories will be added to $\mathcal{D}_0$ for further supervised reasoning tuning, resulting in an improved model with an enhanced reasoning policy.}

\section{Experiments}  
We perform extensive experiments to assess the framework's effectiveness, efficiency, and generalizability in various urban traffic scenarios, addressing the following research questions:
\begin{itemize}[leftmargin=2em]  
\item \textbf{RQ1}: How does CoLLMLight's performance compare to traditional TSC methods and general LLMs?  

\item \textbf{RQ2}: What impacts do CoLLMLight's components have on its performance and efficiency?  

\item \textbf{RQ3}: How do neighboring information and temporal dynamics influence the framework's performance and computational efficiency?
\end{itemize}

\begin{small}  
\begin{table}[t]  
 \small  
 \centering  
 \caption{Statistics of datasets.}  
 \vspace{-5pt}  
 \resizebox{0.48\textwidth}{!}{  
  \begin{tabular}{c|c|c|ccccc}  
   \toprule  
   \multirow{2}{*}{Flow dataset} & \multirow{2}{*}{Structure} & \multirow{2}{*}{\# of vehicles} & \multicolumn{5}{c}{Intersection Connectivity Index} \\
   \cline{4-8}  
   &  & &Max& $Q_3$ &  Std&  Mean& Min \\
   \midrule  
   New York 1 & \multirow{2}{*}{$28\times7$} & 11058 & 2.513 & 0.894 & 0.479 & 0.518 & 0.017 \\
   New York 2 & & 16337 & 3.666 & 1.325 & 0.706 & 0.767 & 0.025 \\
   \midrule  
   Jinan 1 & \multirow{3}{*}{$3\times4$} & 6295 & 0.717 & 0.654 & 0.059 & 0.617 & 0.509 \\
   Jinan 2 & & 4365 & 0.502 & 0.464 & 0.035 & 0.438 & 0.385 \\
   Jinan 3 & & 5494 & 0.622 & 0.573 & 0.048 & 0.544 & 0.48 \\
   \midrule  
   Hangzhou 1 & \multirow{2}{*}{$4\times4$} & 2983 & 0.318 & 0.217 & 0.057 & 0.18 & 0.097 \\
   Hangzhou 2 & & 6984 & 0.815 & 0.492 & 0.17 & 0.413 & 0.217 \\
   \midrule  

   Syn-Train &$4\times4$ & 8000 & 3.463 & 1.785 & 0.979 & 1.38 & 0.028 \\
   \bottomrule  
  \end{tabular}}  
  \label{tab:dataStats}  
\end{table}  
\end{small}  
\subsection{Experimental Setup}
\subsubsection{Datasets}
Our experiments were conducted on seven real-world \cite{wei2019survey} and one synthetic traffic flow dataset. Comprehensive statistics for these datasets are presented in Table \ref{tab:dataStats}. Syn-Train is a dataset that we synthesized. In our experiment, all learning-based methods (RL-based, LLMLight, CoLLMLight) are trained on the Syn-Train and tested for their zero-shot performance in real-world datasets.
\begin{itemize}[leftmargin=2em]  
\item \textbf{Jinan}: A dataset from the Dongfeng sub-district in Jinan, China, consisting of 12 intersections. The dataset includes three traffic flow datasets captured across different time periods, with each intersection featuring two 400-meter roads (east-west) and two 800-meter roads (north-south).  
\item \textbf{Hangzhou}: A dataset from the Gudang sub-district in Hangzhou, China, comprising 16 intersections and two traffic flow datasets across different temporal periods, with each intersection featuring two 800-meter roads (east-west) and two 600-meter roads (north-south). 
\item \textbf{New York}: Collected in Manhattan's Upper East Side using taxi trip data, this extensive dataset encompasses 196 intersections, with each road measuring 300 meters. It includes two large-scale traffic flow datasets from different periods.  
\item \textbf{Syn-Train}: A synthetic dataset comprising 16 intersections, with each road segment measuring 300 meters. This dataset serves as the primary training data for learning-based traffic signal control methods in our experimental framework.  
\end{itemize}
To compare the differences in coordination complexity across these datasets, we define the \textit{Intersection Connectivity Index} as follows:
\begin{definition}[Intersection Connectivity Index (ICI)]  
The Intersection Connectivity Index quantifies an intersection's correlation with its adjacent intersections. An intersection with a high ICI that fails to coordinate with its neighbors may lead to congestion. The ICI for intersection $ \mathcal{V}_i $ can be calculated as follows: 
\begin{equation}  
    ICI =  \sum_{\mathcal{V}_j\in\mathcal{N}(\mathcal{V}_i)}\left(\frac{n_{ij}v}{d_{ij}} \right),  
\end{equation}  
where $\mathcal{N}(\mathcal{V}_i)$ represents the set of neighboring intersections connected to $\mathcal{V}_i$, $v$ is the average vehicle speed, $d_{ij}$ is the road length between $\mathcal{V}_i$ and $\mathcal{V}_j$, and $n_{ij}$ is the average number of vehicles passing between $\mathcal{V}_i$ and $\mathcal{V}_j$ per 30 seconds.  
\end{definition}  
As observed in Table \ref{tab:dataStats}, New York's ICI exhibits significantly higher maximum and upper quartile ($Q_3$) values compared to Jinan and Hangzhou. This suggests that the New York traffic network demands more sophisticated traffic signal cooperation strategies than others.

\subsubsection{Environment Settings}  
We utilize CityFlow \cite{zhang2019cityflow}, a widely used open-source simulator, to perform our experiments. We employ four signal phases as control actions: ETWT (east-west through), ELWL (east-west left-turn), NTST (north-south through), and NLSL (north-south left-turn). Each traffic flow dataset simulates an hour. In the simulation, right-turn movements are allowed at all times. The green signal phase lasts for thirty seconds, followed by a three-second yellow phase and a two-second all-red phase \cite{zhang2022expression, wei2019colight, lai2023large}.

\subsubsection{Compared Methods}  
We compare our proposed method with two conventional transportation methods: FixedTime \cite{koonce2008traffic} and MaxPressure \cite{varaiya2013max}; six RL-based approaches: MPLight \cite{chen2020toward}, AttendLight \cite{oroojlooy2020attendlight}, PressLight \cite{wei2019presslight}, CoLight \cite{wei2019colight}, Efficient-CoLight \cite{wu2021efficient}, and Advanced-CoLight \cite{zhang2022expression}; and a SOTA LLM-based method: LLMLight \cite{lai2023large}. 
Additionally, we assess the performance of general LLMs integrated within our CoLLMLight framework, including Llama 3.1 (8B and 70B), Qwen 2.5 (70B), and the Deepseek-R1 distilled models (R1-8B and R1-32B).

\begin{table*}[htbp]
\caption{Zero-Shot Performance Comparison across Different Datasets (Lower Values Indicate Better Performance). Performance Rankings: best results in boldface, second-best results with \underline{\underline{double underline}}, and third-best results with \underline{single underline}. The lower part of the table shows the performance of various LLMs using the CoLLMLight agent framework.}
\label{tab:performance_comparison}
\centering
\small
\resizebox{\textwidth}{!}{
\begin{tabular}{lcccccccccccccc}
\toprule
\multirow{2}{*}{Models} &
\multicolumn{2}{c}{New York 1} &
\multicolumn{2}{c}{New York 2} &
\multicolumn{2}{c}{Jinan 1} &
\multicolumn{2}{c}{Jinan 2} &
\multicolumn{2}{c}{Jinan 3} &
\multicolumn{2}{c}{Hangzhou 1} &
\multicolumn{2}{c}{Hangzhou 2} \\
& ATT & AWT & ATT & AWT & ATT & AWT & ATT & AWT & ATT & AWT & ATT & AWT & ATT & AWT \\
\hline
FixedTime & 1535.65 & 290.12 & 1771.74 & 428.71 & 481.79 & 70.98 & 441.19 & 66.71 & 450.11 & 69.19 & 616.01 & 73.98 & 486.76 & 72.79 \\
MaxPressure & 1223.26 & 153.27 & 1566.77 & 255.46 & 282.57 & 44.53 & 273.19 & 38.24 & 265.75 & 40.19 & 325.32 & 49.60 & 347.74 & 70.57 \\
MPLight & 1492.15 & 262.23 & 1711.04 & 364.61 & 461.76 & 73.35 & 482.21 & 84.57 & 429.01 & 67.60 & 496.15 & 64.67 & 440.63 & 71.47 \\
AttendLight & 1267.74 & 292.75 & 1755.28 & 496.78 & 392.68 & 69.05 & 310.22 & 63.38 & 337.39 & 67.33 & 327.17 & 72.85 & 348.06 & 68.60 \\
PressLight & 1687.16 & 508.38 & 1894.04 & 489.66 & 376.88 & 87.34 & 410.09 & 139.17 & 367.82 & 93.93 & 599.95 & 259.18 & 539.62 & 317.85 \\
CoLight & 1443.28 & 280.61 & 1704.67 & 392.05 & 516.87 & 93.93 & 474.35 & 79.98 & 491.63 & 86.16 & 526.63 & 74.61 & 479.84 & 88.78 \\
Efficient - CoLight & 1266.04 & 281.32 & 1645.69 & 437.33 & 765.04 & 532.01 & 844.80 & 489.91 & 846.29 & 558.33 & 874.53 & 466.77 & 672.38 & 408.90 \\
Advanced - CoLight & 1037.87 & 185.49 & 1428.90 & 359.79 & 414.25 & 138.18 & 392.89 & 144.74 & 441.79 & 176.21 & 427.99 & 235.86 & 410.79 & 154.06 \\
LLMLight - 8B & 1187.35 & 143.06 & 1599.44 & 388.74 & \textbf{274.97} & 43.98 & 268.57 & 40.45 & \underline{\underline{259.49}} & 40.12 & 311.99 & 39.54 & 332.46 & 61.16 \\
\hline
Llama3.1 - 8B & 1153.74 & 148.92 & 1446.12 & \underline{224.50} & 331.32 & 85.21 & 310.81 & 75.21 & 305.55 & 79.31 & 343.20 & 82.00 & 353.80 & 88.27 \\
Deepseek R1 - 8B & 1096.70 & \underline{120.51} & 1454.74 & 325.28 & 285.94 & 40.85 & 272.83 & 39.51 & 266.74 & 38.92 & 314.19 & 44.75 & 336.20 & 55.06 \\
Deepseek R1 - 32B & \underline{\underline{1011.01}} & \underline{\underline{114.80}} & \underline{\underline{1256.11}} & \underline{\underline{202.76}} & \underline{277.36} & \textbf{35.13} & \underline{\underline{266.98}} & \textbf{33.78} & \underline{260.46} & \underline{34.53} & \underline{305.81} & \textbf{34.18} & \textbf{323.71} & \textbf{39.01} \\
Llama3.1 - 70B & \underline{1023.15} & 143.44 & 1400.00 & 284.74 & 293.09 & 45.56 & 274.78 & 39.09 & 270.74 & 40.21 & 314.08 & 41.98 & \underline{332.06} & 53.65 \\
Qwen2.5 - 72B & 1128.54 & 177.94 & \underline{1392.48} & 244.49 & 279.10 & \underline{35.66} & \underline{267.34} & \underline{\underline{33.81}} & 261.30 & \textbf{34.34} & \underline{\underline{304.57}} & \underline{\underline{34.28}} & 332.43 & \underline{46.82} \\
\textbf{CoLLMLight - 8B} & \textbf{920.98} & \textbf{85.90} & \textbf{1218.94} & \textbf{170.13} & \underline{\underline{276.41}} & \underline{\underline{35.25}} & \textbf{266.89} & \underline{34.50} & \textbf{258.89} & \underline{\underline{34.47}} & \textbf{303.77} & \underline{34.72} & \underline{\underline{326.02}} & \underline{\underline{44.12}} \\

\bottomrule
\end{tabular}
}
\label{tab:perf}
\end{table*}

\subsubsection{Evaluation Metrics}  
We adopt \textsl{Average Travel Time} (ATT) and \textsl{Average Waiting Time} (AWT) \cite{wei2021recent, zhang2022expression, lai2023large} as evaluation metrics to evaluate the performance of various traffic signal control methods. Specifically, ATT quantifies the average duration required for vehicles to travel from their origins to their respective destinations. AWT quantifies the average queuing time experienced by vehicles at each intersection within the road network.

    
    

\subsection{Implementation Details}  
We conducted our experiments on Linux servers that were equipped with two A800 GPUs. All RL methods are trained with consistent hyperparameters: a learning rate of $1 \times 10^{-3}$, a replay buffer size of 12,000, a sample size of 3,000, and a hidden size of 20. For LLMs, the temperature parameter is set to 0.1. We perform LoRA fine-tuning on Llama 3.1 (8B) to obtain LLMLight and CoLLMLight, maintaining LoRA settings with a rank of 8 and a scaling factor $\alpha$ of 16, along with a learning rate of $1 \times 10^{-4}$. During the reasoning tuning stage, we construct 2,802 trajectories and sample an additional 552 trajectories from Syn-Train during the policy refinement stage.

\subsection{Comparative Performance (RQ1)}   
In the first experiment, all learning-based methods were trained on the synthetic traffic flow dataset, Syn-Train, and evaluated on seven real-world traffic flow datasets to assess their zero-shot performance. The results are presented in Table \ref{tab:perf}.   

Specifically, CoLLMLight consistently achieves SOTA or comparable performance against all baselines, demonstrating superior generalization and cooperation capabilities. This indicates that CoLLMLight can effectively adapt its learned policies to various traffic scenarios. Among the previous methods, MaxPressure demonstrates superior generalization capabilities compared to most RL-based approaches, which often struggle to adapt their performance to new scenarios due to their data-centric nature. While LLMLight excels in Jinan 1, it struggles with complex coordination in New York, revealing the limitations of its localized decision-making.

In our evaluation of generalist LLMs within the CoLLMLight framework, Deepseek R1-32B exhibits performance comparable to CoLLMLight-8B in Jinan and Hangzhou. However, in the New York dataset, which presents higher coordination complexity, CoLLMLight-8B outperforms all baselines, achieving a 25.17\% reduction in AWT for New York 1 and a 16.09\% reduction for New York 2 compared to the second-best performance. 

\begin{figure}[t] 
    \centering  
    \includegraphics[width=\linewidth]{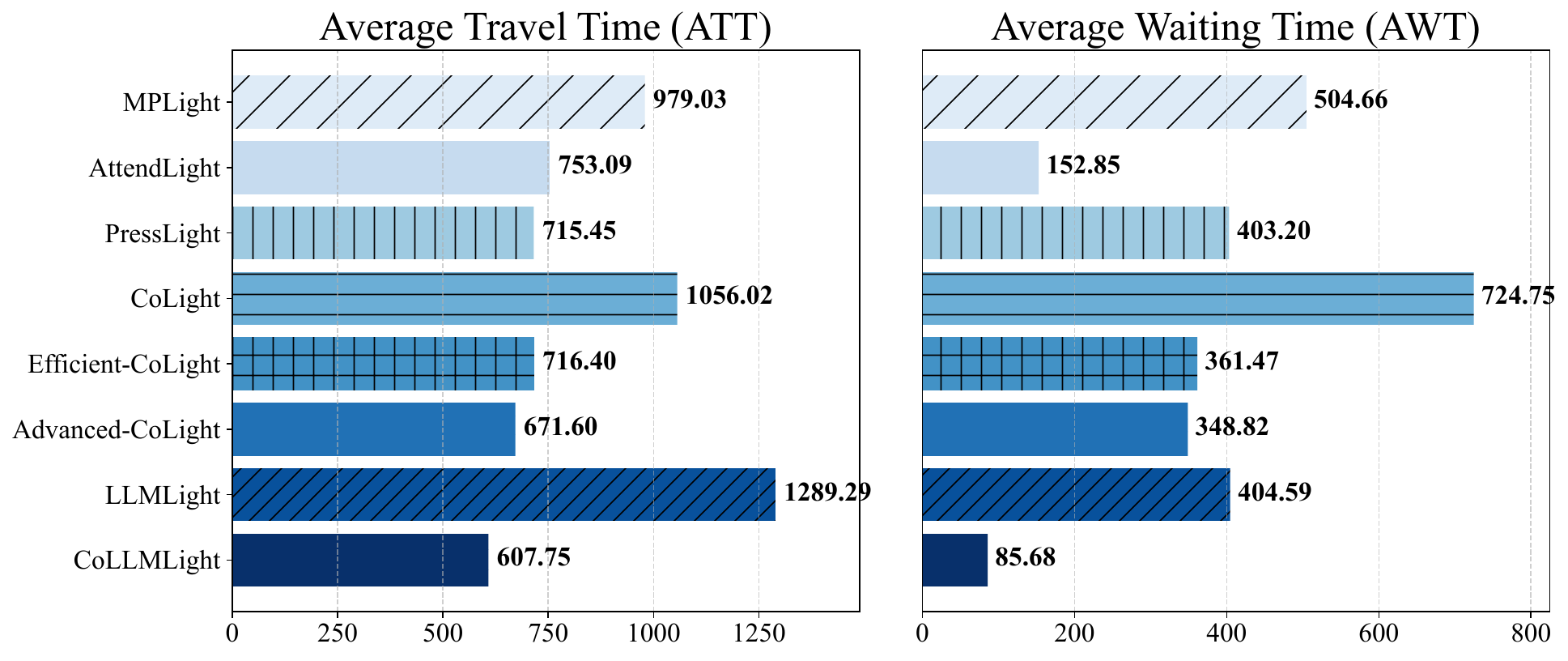} 
    \caption{Comparative Performance of Learning-based Methods at Syn-Train}  
    \label{fig:syn_train_performance}  
\end{figure}  

We also report the performance of learning-based methods on the Syn-Train dataset in Figure \ref{fig:syn_train_performance}. CoLLMLight outperforms all other methods in this dataset due to its effective policy refinement, which enhances its cooperative abilities within the traffic environment, leading to superior TSC performance. In contrast, LLMLight demonstrates poor performance due to its policy not considering coordination among intersections, which is critical at Syn-Train. Among RL-based methods, Advanced-CoLight shows the best learning performance in the Syn-Train dataset; however, it struggles to transfer this performance to real-world datasets. In contrast, CoLLMLight generalizes the policy learned from Syn-Train, achieving effective coordination and demonstrating its robustness across different traffic scenarios. Additionally, we compare DeepSeek-R1 (671B) and O1mini on the New York 1 dataset, as detailed in the Appendix \ref{sec:pefr1}. The results show that CoLLMLight outperforms the SOTA LLMs within complex coordination scenarios.


\subsection{Ablation Study (RQ2)}

\subsubsection{Ablation of Reasoning Strategy }
The complexity-aware reasoning mechanism balances performance and efficiency by intelligently identifying current cooperation complexity and switching among three reasoning strategies: $f^{\text{no-coop}}_{\text{LLM}}$, $f^{\text{simple}}_{\text{LLM}}$, and $f^{\text{complex}}_{\text{LLM}}$. These strategies are denoted as \textbf{NC}, \textbf{SR}, and \textbf{CR}. This section evaluates the impact of these reasoning strategies by comparing the performance and total inference time across various variants.

(1) \textbf{Single Reasoning Strategy}. These variants do not incorporate complexity identification and rely solely on a single reasoning strategy to determine the optimal traffic signal. 

\begin{figure}[t] 
    \centering  
    \includegraphics[width=\linewidth]{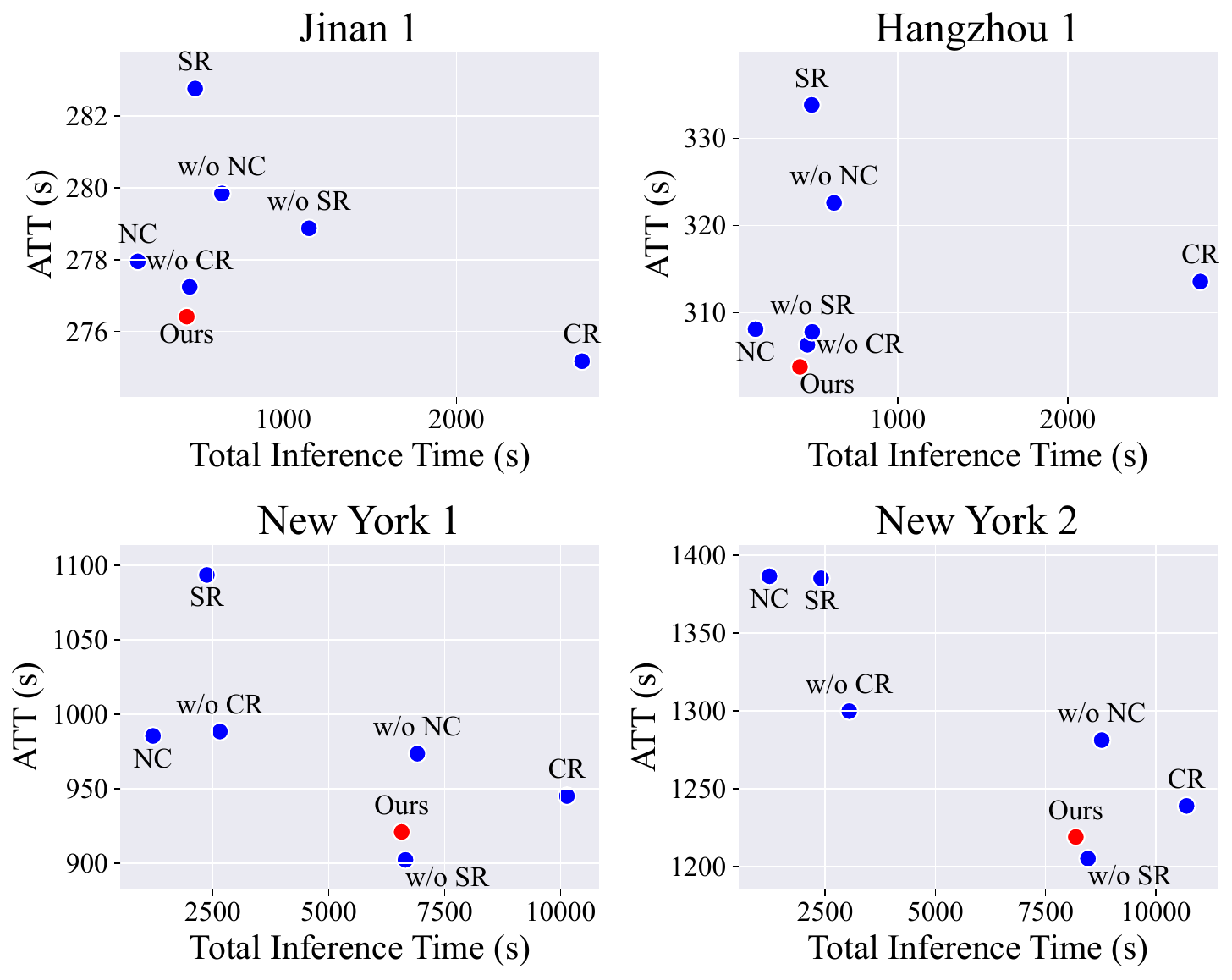} 
    \caption{Performance and efficiency of various variants across four datasets. Methods positioned towards the lower left ($\swarrow$) exhibit superior performance ($\downarrow$) while incurring lower time costs ($\leftarrow$).}  
    \label{fig:ablation_reasoning}  
\end{figure}

(2) \textbf{Two Reasoning Strategies} (\textbf{w/o NC, w/o SR, w/o CR}). These variants integrate two reasoning strategies. For instance, w/o NC indicates the CoLLMLight without $f^{\text{no-coop}}_{\text{LLM}}$, but with $f^{\text{simple}}_{\text{LLM}}$ and $f^{\text{complex}}_{\text{LLM}}$. The experimental results are presented in Figure \ref{fig:ablation_reasoning}. 

Among these three variants, w/o SR demonstrates strong cost-effectiveness in the New York dataset but fails in the Jinan and Hangzhou datasets due to its low efficiency in simple cooperation scenarios. In contrast, our method integrates all three strategies to address the cooperation problem more flexibly. By utilizing reasoning strategies tailored to different traffic scenarios, our method outperforms the complex reasoning in most datasets while achieving significant efficiency improvements.

\begin{table}[t]
\caption{Abalation Tests of Optimization Stages}
\label{tab:ablation_model_optimization_results}
\centering
\small
\scalebox{0.9}{
\begin{tabular}{lccccccc}
\toprule
\multirow{2}{*}{Model} & \multicolumn{3}{c}{Jinan} & \multicolumn{2}{c}{Hangzhou} & \multicolumn{2}{c}{New York} \\
\cmidrule(lr){2-4} \cmidrule(lr){5-6} \cmidrule(lr){7-8}
 & 1 & 2 & 3 & 1 & 2 & 1 & 2 \\
\midrule
w/o Both & 331.32 & 310.81 & 305.55 & 343.20 & 353.80 & 1153.74 & 1446.12 \\
w/o RO   & 312.39 & 294.47 & 291.66 & 338.39 & 351.85 & 1167.37 & 1500.65  \\
w/o PR   & 278.35 & 267.73 & 261.03 & 304.52 & 327.92 & 1085.88 & 1368.03 \\
Ours     &\textbf{ 276.41} & \textbf{266.89 }& \textbf{258.89} & \textbf{303.77 }& \textbf{326.02} &\textbf{ 920.98}  & \textbf{1218.94} \\
\bottomrule
\end{tabular}
}
\end{table}

\subsubsection{Ablation of Optimization Stages}
This ablation study evaluates the effectiveness of our training stages by comparing our model against three variants: (1) \textbf{w/o Both}: the base Llama3.1 8b model, which serves as the initial baseline without any specialized training; (2) \textbf{w/o RO}: a variant that omits reasoning chain optimization, which directly applies the base Llama3.1 8b model to interact with the environment and collect trajectories for policy refinement; (3) \textbf{w/o PR}: a variant that has conducted reasoning chain optimization but discards policy refinement. The comparative results are shown in Table \ref{tab:ablation_model_optimization_results}.

The experimental results show that the ablation of any components leads to a drop in performance, highlighting the critical roles of reasoning chain optimization and policy refinement in enhancing our model's performance. Without these optimizations, the base model struggles to produce effective traffic signal control trajectories, especially in complex urban traffic scenarios, such as those in New York. Moreover, if we directly apply refinement using environmental feedback, performance in New York may actually decline, as indicated by the results of the w/o RO variant. This decline is due to the lack of a fundamental reasoning process that is essential for effective control strategies in various urban environments.

The results of the w/o PR variant indicate the model's fundamental reasoning capabilities for cooperative traffic signal control has been activated, showing strong performance in smaller traffic networks such as Jinan and Hangzhou, where traffic patterns are relatively simpler. However, without further refining its policy with environmental feedback, the model still struggles in more complex environments like New York, which features intricate traffic dynamics and higher coordination requirements. By further fine-tuning the model's policy, our method demonstrates substantial performance improvements, particularly in the New York dataset. These gains illustrate that feedback refinement effectively enhances the model's ability to generalize across diverse and challenging traffic environments, bridging the gap between learned imitation and optimal control strategies. 

\begin{figure}[H]   
    \centering  
    \includegraphics[width=\linewidth]{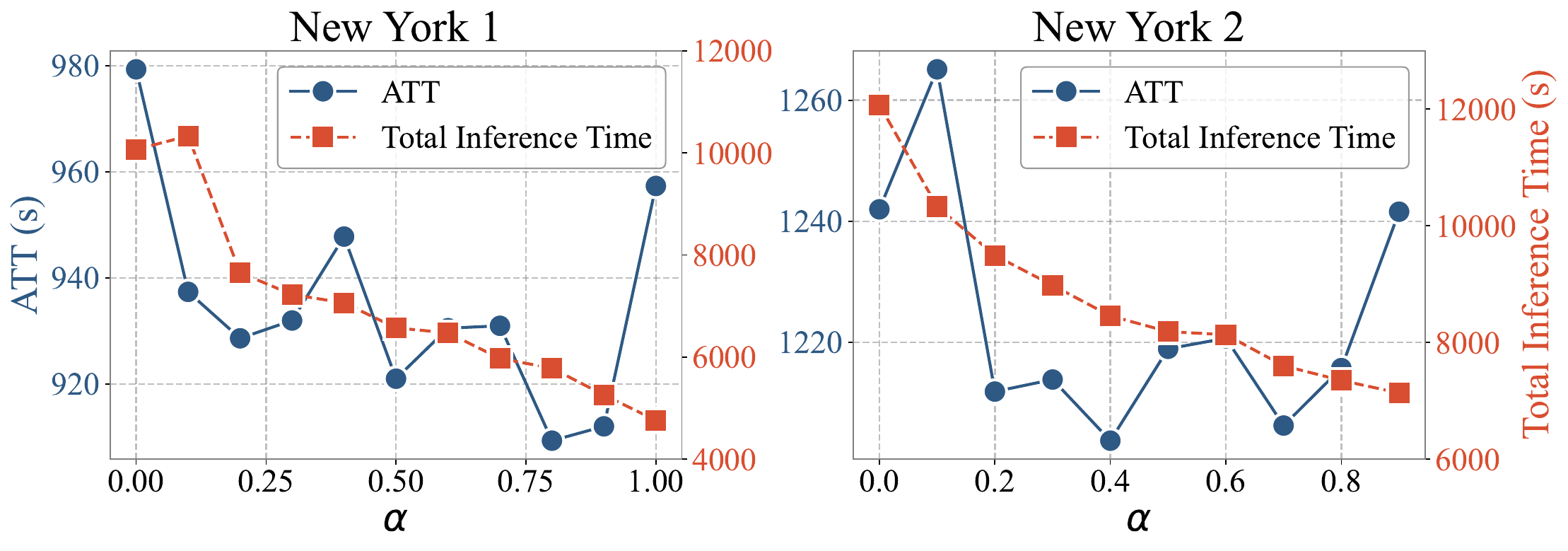}  
    \caption{Average Travel Time and Total Inference Time with varying communication threshold $\alpha$.}  
    \label{fig:communication_threshold}  
\end{figure}  

\subsection{Analysis of Spatiotemporal Information (RQ3)}  
In this section, we analyze the impact of the spatiotemporal information processed by our agent by conducting a sensitivity analysis on two key hyperparameters (the communication threshold $\alpha$ and the historical window size $\Delta t$) to adjust neighboring information and temporal information, respectively.
\subsubsection{Analysis of Neighboring Information}  
In our observation collection process, only neighboring lanes with occupancy above the threshold $\alpha$ are aggregated. A larger $\alpha$ results in less neighboring information being considered. We vary $\alpha$ from 0 to 1 while keeping $\Delta t = 5$, and the results are reported in Figure \ref{fig:communication_threshold}.  

Specifically, when we set $\alpha = 0$, the intersection must consider all neighboring lanes, leading to longer processing times and a higher likelihood of errors. As $\alpha$ increases, unimportant lanes with low occupancy are filtered out, reducing inference time and allowing the model to focus on the critical lanes that may contribute to congestion, thereby improving performance. However, if $\alpha$ becomes too high, critical lanes may also be disregarded, resulting in decreased performance. In our main experiments, we set $\alpha = 0.5$.

\begin{figure}[H]   
    \centering  
    \includegraphics[width=\linewidth]{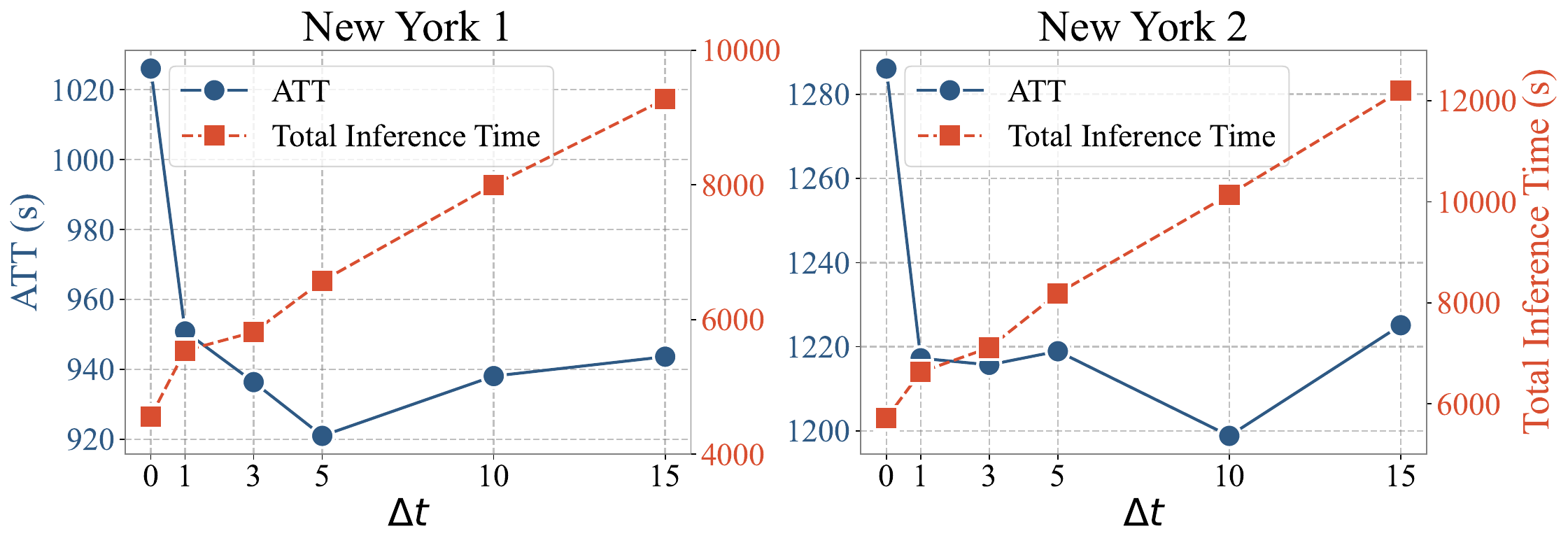}  
    \caption{Average Travel Time and Total Inference Time with varying historical window size $\Delta t$.}  
    \label{fig:historical_window_size}  
\end{figure} 

\subsubsection{Analysis of Historical Information $\Delta t$} 
The value of $\Delta t$ determines the amount of historical information available to the agent. We vary $\Delta t$ within the range $[0, 1, 3, 5, 10, 15]$ while keeping $\alpha = 0.5$. The results are reported in Figure \ref{fig:historical_window_size}.  

Specifically, when $\Delta t = 0$, the agent does not consider any historical information in its decision-making, resulting in the lowest performance across all datasets. Once the agent considers a timestep of information, the ATT shows a significant decrease, highlighting the importance of historical information for effective traffic signal control. However, the computational overhead increases linearly with $\Delta t$ due to the larger input text processed by the LLM agent. Importantly, performance does not consistently improve with an increased historical window size. We observe performance degradation when $\Delta t$ exceeds 5 or 10 timesteps. This decline can be attributed to the increased complexity and potential noise introduced by longer input contexts. To balance performance and efficiency, we set $\Delta t = 5$ for our main experiments.

\section{Related Work}

\subsection{Traffic Signal Control}  
Traffic signal control presents a significant challenge within intelligent transportation systems, with methodological approaches evolving across traditional transportation methods, RL-based methods, and LLM TSC agents. Traditional strategies include FixedTime \cite{koonce2008traffic}, which relies on predetermined cycle lengths and phase allocations, and Maxpressure \cite{varaiya2013max}, which is designed to mitigate over-saturation by minimizing the difference in total queue lengths between incoming and outgoing traffic flows. RL-based methods have transformed the field through innovative neural architectures and advanced feature representations. Notable approaches include FRAP \cite{zheng2019learning}, which introduced a dynamic phase-level interaction network;  CoLight \cite{wei2019colight}, leveraging graph attention networks \cite{velivckovic2017graph} for inter-intersection coordination; CosLight \cite{ruan2024coslight}, utilizing MLP to construct the collaborator matrix among all intersections. Concurrently, researchers have focused on developing more sophisticated traffic state representations. PressLight \cite{wei2019presslight} pioneered pressure-based state representations to enhance traffic management, Efficient-XLight \cite{wu2021efficient} introduced efficient pressure, and Advanced-XLight \cite{zhang2022expression} further advanced performance by combining it with effective vehicle count. 

Recently, emerging LLMs have demonstrated transformative potential, as seen in frameworks like LLMLight \cite{lai2023large}, which showcases exceptional generalization by mimicking human-like reasoning processes. However, current LLM-based TSC agents ignore the coordination among intersections, which is crucial for alleviating traffic congestion. 
 
\subsection{Multi-Agent Cooperation of Large Language Models}  
Recently, LLM-based multi-agent systems have achieved considerable progress across various domains \cite{ijcai2024p890}, demonstrating remarkable performance in software development \cite{hongmetagpt, qian2024chatdev}, multi-robot systems \cite{mandi2024roco, zhang2024building}, and societal simulation \cite{park2023generative}. These innovative systems leverage multiple autonomous agents to collaboratively engage in complex planning, nuanced discussions, and sophisticated decision-making, effectively mimicking the intricate cooperative dynamics of human group interactions. For instance, MetaGPT \cite{hongmetagpt} embeds human workflow processes into language model agents, utilizing an assembly line strategy that assigns specialized roles to different agents to facilitate seamless cooperation. CoMAL \cite{yao2024comal} introduces a collaboration module that enables autonomous vehicles to collectively discuss and strategize, dynamically allocating roles to optimize collaborative performance. 

However, many multi-agent LLM frameworks rely on iterative inter-agent communication, requiring multiple LLM calls for discussions and consensus-building, which limits their ability to provide the real-time responsiveness essential for traffic signal control. In contrast, CoLLMLight is an efficient cooperative agent that independently analyzes the dynamics of its collaborators, considers their interests, and makes decisions aimed at optimizing global traffic.

\section{Conclusion}  
This paper introduces CoLLMLight, a cooperative LLM agent framework specifically designed for network-wide traffic signal control. To effectively address the complexities of traffic interactions and foster cooperation among intersections, we model the spatiotemporal information that LLM agent can receive in the traffic environment and propose a Spatiotemporal-Aware Cooperative Decision-Making approach to comprehensively analyze current traffic conditions, predict future states under various signal configurations, evaluate their network-wide impacts, and select optimal signals to enhance overall traffic flow.  
To promote efficiency, we introduce a Complexity-Aware Reasoning mechanism that classifies traffic scenario complexity into three distinct levels and integrates two additional efficient decision-making strategies for simpler cooperative scenarios.  
Additionally, we propose a Simulation-Driven Fine-Tuning approach to optimize our agent's reasoning chain and refine its policy through interaction with the environment. Extensive experiments conducted on both synthetic and real-world datasets validate the superior performance of our framework, demonstrating its capability to effectively manage complex urban traffic scenarios with remarkable efficiency. 
In the future, we plan to investigate a hierarchical asynchronous Multi-Agent LLM framework for traffic signal control, aimed at guiding the real-time decision-making of intersection agents through regional-level planning.

\bibliographystyle{ACM-Reference-Format}
\bibliography{sample-base}

\newpage
\appendix

\section{Appendix}
\subsection{Settings of Traffic Signal Control}
\label{sec:tsc_setting}

\begin{figure}[H]   
    \centering  
    \includegraphics[width=\linewidth]{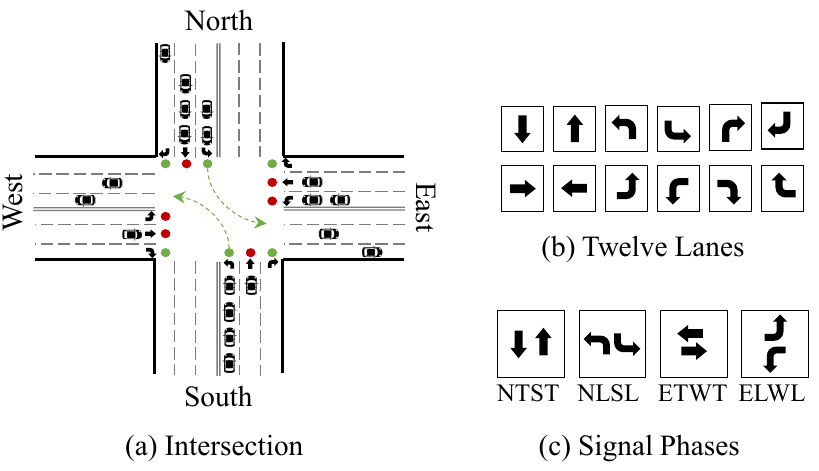}  
    \caption{An illustration of intersections, lanes, and signal phases. The signal phases include ETWT (east-west through movement), ELWL (left-turn from east and west), NTST (north-south through movement), and NLSL (left-turn from north and south)}  
    \label{fig:tsc_setting}  
\end{figure}

We present the most used settings in Figure \ref{fig:tsc_setting}.

\subsection{Prompt Template and Decision Making Process}
\begin{figure*}[htbp]  
    \centering  
    \includegraphics[width=0.9\textwidth]{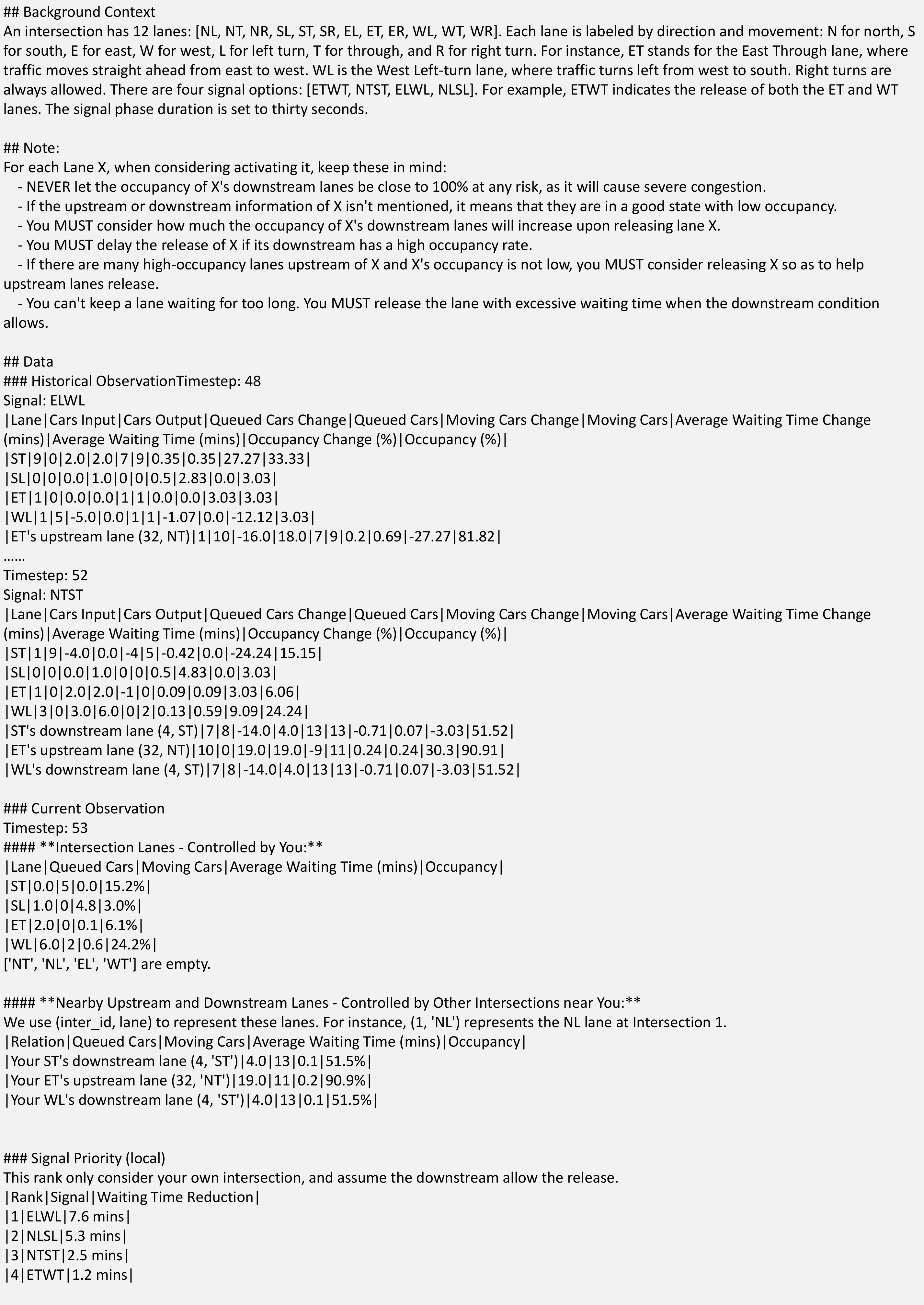}  
    \caption{Prompt Template Part 1}  
    \label{fig:prompt1}  
\end{figure*}

\begin{figure*}[htbp]  
    \centering  
    \includegraphics[width=0.9\textwidth]{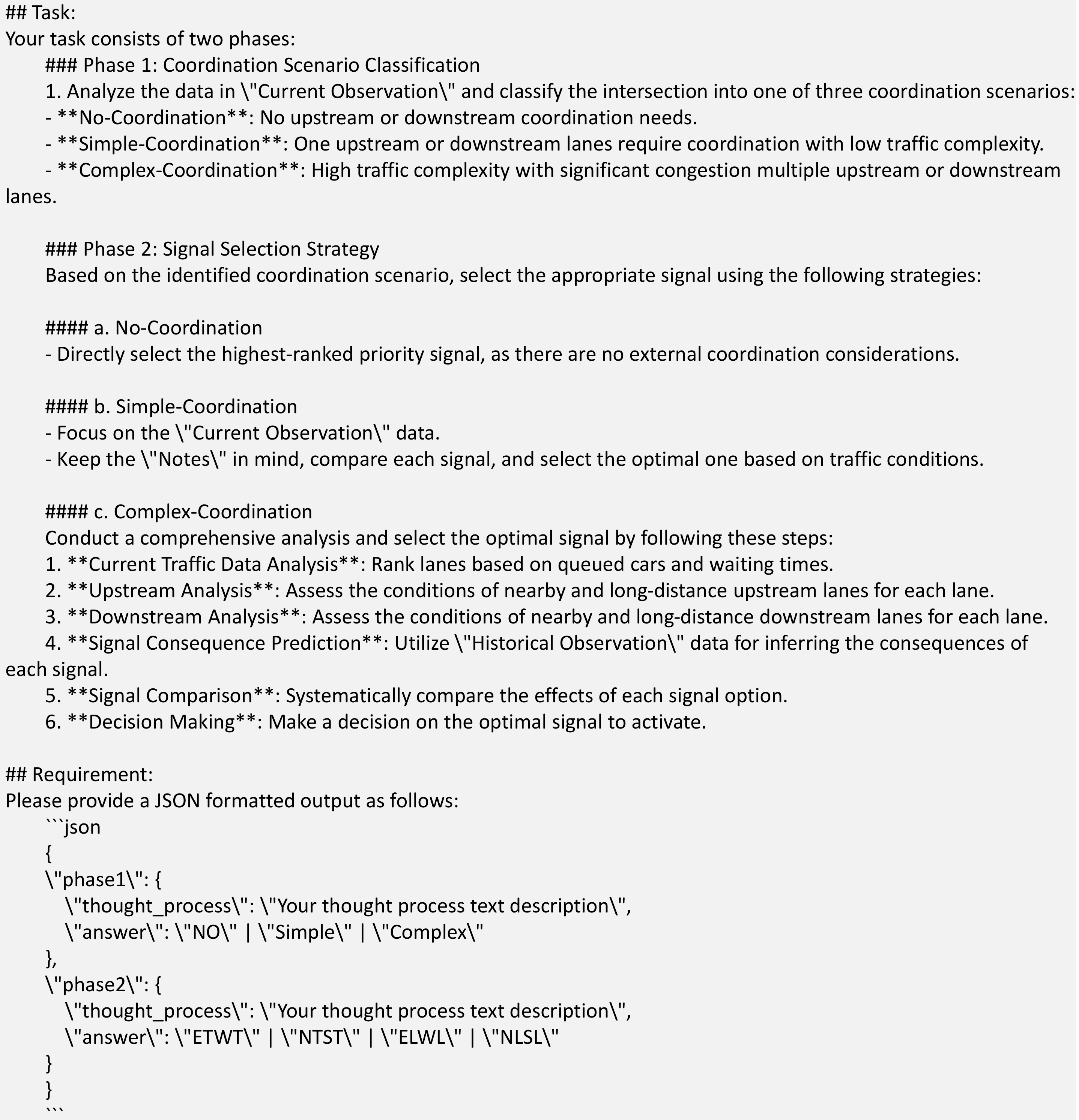}  
    \caption{Prompt Template Part 2}  
    \label{fig:prompt2}  
\end{figure*}

\begin{figure*}[htbp]  
    \centering  
    \includegraphics[width=0.9\textwidth]{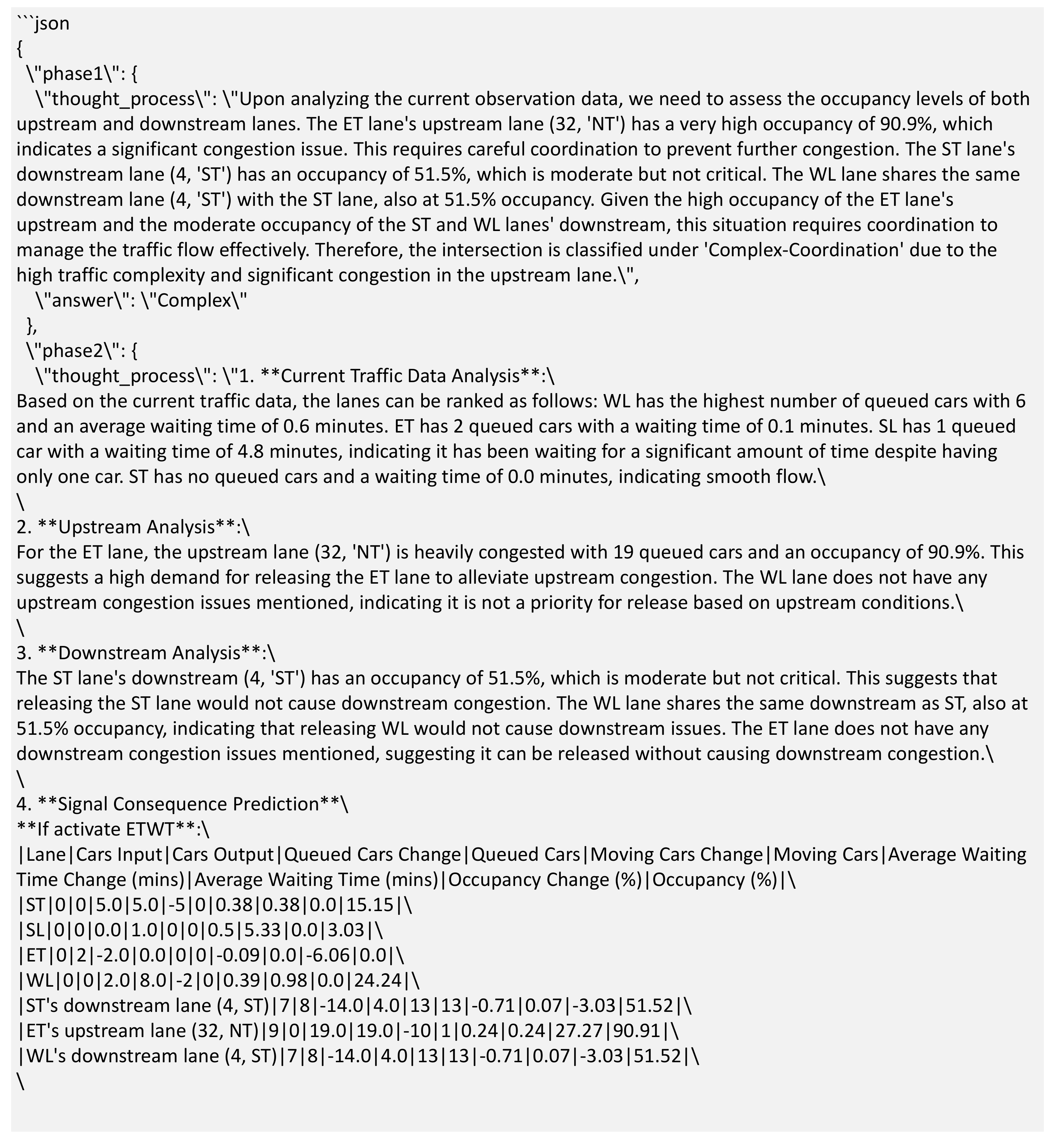}  
    \caption{Decision Making Process 1}  
    \label{fig:dm1}  
\end{figure*}

\begin{figure*}[htbp]  
    \centering  
    \includegraphics[width=0.9\textwidth]{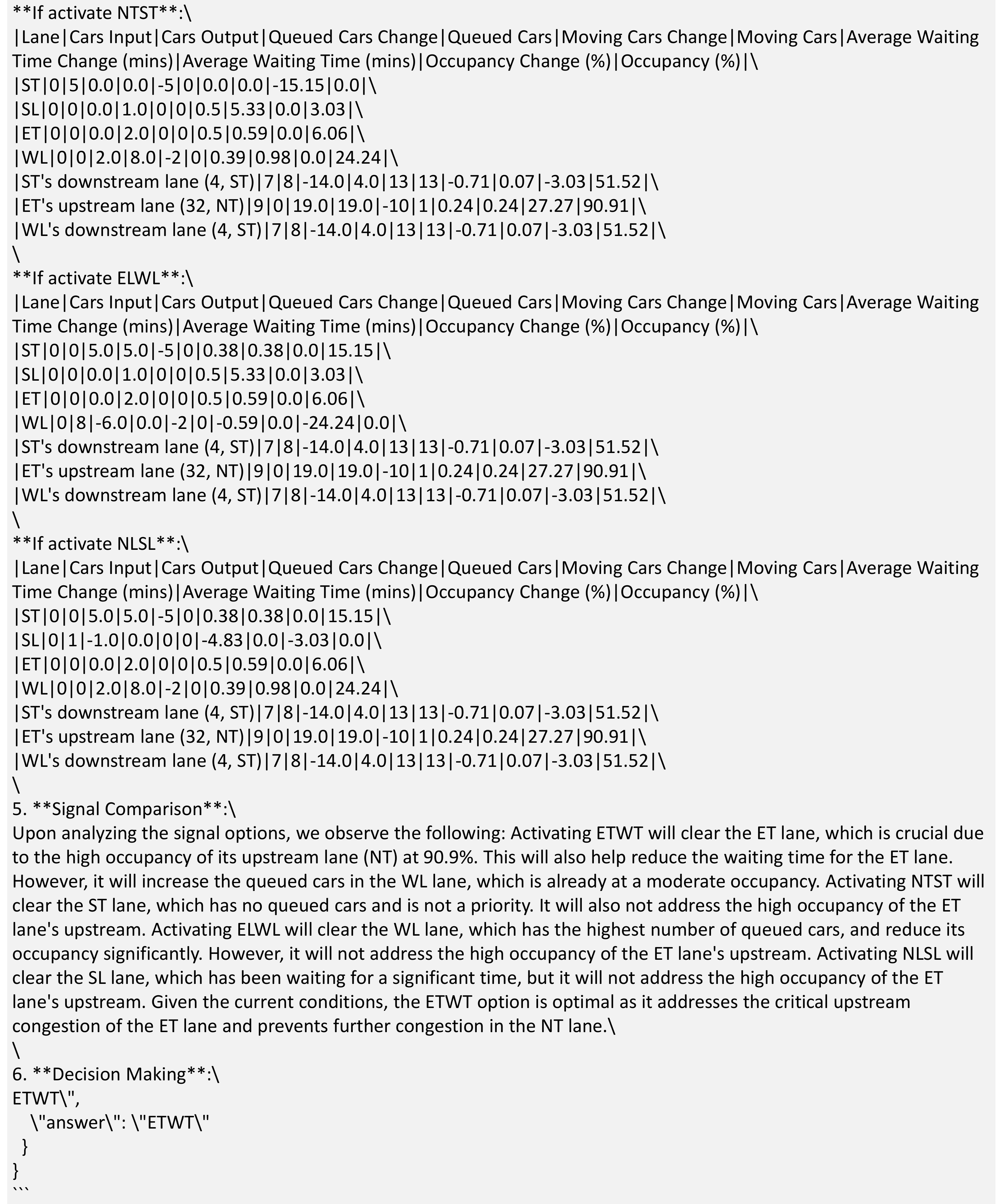}  
    \caption{Decision Making Process 2}  
    \label{fig:md2}  
\end{figure*}

We present our prompt template in Figures \ref{fig:prompt1} and \ref{fig:prompt2}, and the decision-making process in Figures \ref{fig:dm1} and \ref{fig:md2}.
\label{sec:prompt_template}

\subsection{Compared Methods}
\begin{itemize} 
    \item \textbf{FixedTime} \cite{koonce2008traffic}: A traditional signal control method using static, predefined cycle lengths and phase times.  
    \item \textbf{MaxPressure} \cite{varaiya2013max}: A SOTA transportation method selecting signal phases based on queue length pressure between upstream and downstream intersections. 
    \item \textbf{MPLight} \cite{chen2020toward}: It utilizes pressure as both observation and reward, based on the FRAP. 
    \item \textbf{AttendLight} \cite{oroojlooy2020attendlight}: It uses attention mechanisms to predict phase transitions and construct observation features.  
    \item \textbf{PressLight} \cite{wei2019presslight}: Applies Deep Reinforcement Learning (DRL) to optimize intersection pressure.  
    \item \textbf{CoLight} \cite{wei2019colight}: Employs graph attention network (GAT) for inter-intersection communication.  
    \item \textbf{Efficient-CoLight} \cite{wu2021efficient}: Enhances CoLight by integrating efficient pressure observations.  
    \item \textbf{Advanced-CoLight} \cite{zhang2022expression}: A SOTA MARL-based method that enhances CoLight by integrating advanced traffic state features, including pressure and effective running vehicles. 
    \item \textbf{LLMLight} \cite{lai2023large}: A SOTA LLM TSC agent that employs a reasoning process mimicking human-like intuition to optimize traffic management.  
\end{itemize}
\subsection{Performance Comparison with DeepSeek R1-671B and O1-mini}
\label{sec:pefr1}
We present the performance comparison results of CoLLMLight against the SOTA LLMs, DeepSeek R1-671B, and O1-mini, in Tables \ref{tab:performance_10_20} and \ref{tab:performance_100_60}, respectively.  

For DeepSeek R1-671B, we conducted an experiment using the New York 1 dataset, selecting the 10\% most congested intersections to be controlled by the LLM agent, and tested its performance over a simulation time span of 1200 seconds. For O1-mini, we evaluated its performance in a complete setting, controlling all 196 intersections in the New York 1 dataset and simulating traffic for 3600 seconds. The results further demonstrate the superior performance of CoLLMLight.

\begin{table}[h]  
    \centering  
    \caption{Performance Comparison between R1 and CoLLMLight, in a setting of 10\% controlled by LLM Agents and 1200 seconds simulation.}  
    \begin{tabular}{@{}lcc@{}}  
        \toprule  
        \textbf{Model} & \textbf{AWT} & \textbf{ATT} \\ \midrule  
        Deepseek R1-671B & 97.148 & 846.587 \\   
        CoLLMLight-8B & \textbf{70.144} & \textbf{834.433} \\   
        \bottomrule  
    \end{tabular}  
    \label{tab:performance_100_60}  
\end{table}  

\begin{table}[h]  
    \centering  
    \caption{Performance Comparison between O1mini and CoLLMLight, in a setting of 100\% controlled by LLM Agents and 3600 seconds simulation.}  
    \begin{tabular}{@{}lcc@{}}  
        \toprule  
        \textbf{Model} & \textbf{AWT} & \textbf{ATT} \\ \midrule  
        O1mini & 95.20 & 986.91 \\   
        CoLLMLight-8B & \textbf{85.90} & \textbf{920.98} \\   
        \bottomrule  
    \end{tabular}  
    \label{tab:performance_10_20}  
\end{table}








\end{document}